# From Shannon's Channel to Semantic Channel via New Bayes' Formulas for Machine Learning

Chenguang Lu

lcguang@foxmail.com

**Abstract.** A group of transition probability functions form a Shannon's channel whereas a group of truth functions form a semantic channel. By the third kind of Bayes' theorem, we can directly convert a Shannon's channel into an optimized semantic channel. When a sample is not big enough, we can use a truth function with parameters to produce the likelihood function, then train the truth function by the conditional sampling distribution. The third kind of Bayes' theorem is proved. A semantic information theory is simply introduced. The semantic information measure reflects Popper's hypothesis-testing thought. The Semantic Information Method (SIM) adheres to maximum semantic information criterion which is compatible with maximum likelihood criterion and Regularized Least Squares criterion. It supports Wittgenstein's view: the meaning of a word lies in its use. Letting the two channels mutually match, we obtain the Channels' Matching (CM) algorithm for machine learning. The CM algorithm is used to explain the evolution of the semantic meaning of natural language, such as "Old age". The semantic channel for medical tests and the confirmation measures of test-positive and test-negative are discussed. The applications of the CM algorithm to semi-supervised learning and non-supervised learning are simply introduced. As a predictive model, the semantic channel fits variable sources and hence can overcome class-imbalance problem. The SIM strictly distinguishes statistical probability and logical probability and uses both at the same time. This method is compatible with the thoughts of Bayes, Fisher, Shannon, Zadeh, Tarski, Davidson, Wittgenstein, and Popper. It is a competitive alternative to Bayesian inference.

**Keywords:** Shannon's channel, Semantic channel, Bayes' theorem, Truth function, Multi-label classification, Semi-supervised learning, Mixture models.

## 1 Introduction

According to Tarski's theory of truth [1] and Davidson's truth-conditional semantics [2], the semantic meaning of a hypothesis or label is ascertained by its truth function. According to Wittgenstein's view, the meaning of a word lies in its use [3]. Can we obtain the truth function of a word from its use?

In natural language processing and machine learning, we need to obtain the meanings or truth functions of some labels according a sample with the labels, which can be considered as a group of ostensive definitions [3]. Although multi-label learning has obtained remarkable successes [4], it is still not easy to get the truth functions of labels especially when the semantic meaning is fuzzy. This study is to obtain the truth functions from samples by the third kind of Bayes' theorem and the semantic information method. Further, it uses the truth function as a new tool for machine learning from larger samples. So, we use sampling distributions instead of sequences of sample points to train predictive models.

After Shannon founded the information theory in1948 [5], Bar-Hillel and Carnap proposed a semantic information measure in 1952 [6]. Lately, others proposed some different measures and theories [7-9]. However, these studies do not combine semantic information with hypothesis-testing. In recent two decades, cross entropy method used has shown its power for hypothesis-testing [10]. Actually, Earlier in 1991, Lu defined the semantic mutual information with average log(normalized likelihood), which was the mutual cross entropy [11]. Lu used the truth function to produce the predicted probability (i.e. semantic likelihood) and set up a semantic information theory [12. 13]. Recently, we found that this theory plus a pair of new Bayes' formulas could be used to improve hypothesis-testing and machine learning including semi-supervised learning [14] and unsupervised leaning [15]. Basing on Lu's works and our recent studies, this paper tries to propose a new mathematical frame with semantic channel and the third kind of Bayes' theorem (newly proposed) for machine learning.

In the following, we successively introduce the mathematical methods (including Bayes' Theorem III, Semantic channel, semantic likelihood), the semantic information theory, and the Channels' Matching algorithm or CM algorithm for machine learning. The last part is a summary with some discussions.

## 2 Mathematical Methods

### 2.1 Distinguishing Statistical probability and Logical Probability

**Definition 2.1.1** Let $U$ denote the instance set and $X$ denote the discrete random variable taking a value from $U$. That means $X \in U=\{x_1, x_2, …\}$. For theoretical convenience, we assume that $U$ is one-dimensional. Let $V$ denote the set of selectable hypotheses or labels and $Y \in V=\{y_1, y_2, …\}$.

**Definition 2.1.2** A hypothesis $y_j$ is also a predicate $y_j(X)=$ "$X \in A_j$". For each $y_j$, $U$ has a subset $A_j$, every instance in which makes $y_j$ true. Let $P(Y=y_j)$ denote the statistical probability of $y_j$, and $P(X \in A_j)$ denote the Logical Probability (LP) of $y_j$. For simplicity, let $P(y_j)= P(Y=y_j)$ and $T(y_j)=T(A_j)= P(X \in A_j)$.

We call $P(X \in A_j)$ the LP because according to Tarski's theory of truth [14], $P(X \in A_j)=P($"$X \in A_j$" is true$)=P(y_j$ is true$)$. Hence the conditional LP of $y_j$ for given $X$ is the feature function of $A_j$ or the truth function of $y_j$. Let it be denoted by $T(A_j|X)$. $T(A_j|X)$ ascertains the semantic meaning of $y_j$. There is

$$T(A_j) = \sum_i P(x_i)T(A_j | x_i) \qquad (2.1)$$

Note that statistical probability distribution, such as $P(Y)$, $P(Y|x_i)$, $P(X)$, and $P(X|y_j)$, are normalized whereas the LP distribution is not normalized. For example, generally,

$$T(A_1)+T(A_2)+\ldots+T(A_n) \geqslant 1,\ T(A_1|x_i)+T(A_2|x_i)+\ldots+T(A_n|x_i) \geqslant 1$$

$P(A_j)=T(A_j)$ only when $\{A_1, A_2, \ldots, A_n\}$ is a partition of $U$ and $Y$ is always correctly used. $T(A_j|X)$ is similar to $P(y_j|X)$; yet, its maximum is 1.

When the sets are fuzzy [16], we use $\theta_j$ to denote a fuzzy set, which is also a predictive model or the sub-model of $\theta$. Then the truth function of $y_j$ or the membership function of $\theta_j$ becomes $T(\theta_j|X)$. The likelihood function $P(X|\theta_j)$ in this paper is equal to $P(X|\theta, y_j)$ in popular methods. The usage of the sub-model $\theta_j$ will make the predictive model flexile and the statements clearer.

### 2.2 Three Kinds of Bayes' Theorems

The Bayes' theorem is described by Bayes' formulas. Actually, this theorem has three forms, which are used by Bayes [17], Shannon [5], and this paper respectively.

**Bayes' Theorem I** (used by Bayes): Assume that sets $A, B \in 2^U$, $A^c$ is the complementary set of $A$, $T(A)=P(X \in A)$, and $T(B)= T(A)=P(X \in B)$. Then

$$T(B|A)=T(A|B)T(B)/T(A),\ T(A)= T(A|B)T(B)+ T(A|B^c)T(B^c) \qquad (2.2)$$

$$T(A|B)=T(B|A)T(A)/T(B),\ T(B)= T(B|A)T(A)+ T(B|A^c)T(A^c) \qquad (2.3)$$

Note there is only one random variable $X$ and two logical probabilities.

**Bayes' Theorem II** (used by Shannon): Assume that $X \in U$, $Y \in V$, $P(x_i)=P(X=x_i)$, and $P(y_j)= P(Y=y_j)$. Then

$$P(x_i | y_j) = P(y_j | x_i)P(x_i) / P(y_j),\ P(y_j) = \sum_i P(x_i)P(y_j | x_i) \qquad (2.4)$$

$$P(y_j | x_i) = P(x_i | y_j)P(y_j) / P(x_i),\ P(x_i) = \sum_j P(y_j)P(x_j | y_j) \qquad (2.5)$$

Note there are two random variables and two statistical probabilities. In each of the above two theorems, two formulas are symmetrical and denominators are normalizing constants.

**Bayes' Theorem III:** Assume that $P(X)=P(X=$any$)$ and $T(A_j)=P(X \in A_j)$. Then

$$P(X | A_j) = T(A_j | X)P(X) / T(A_j),\ T(A_j) = \sum_i P(x_i)T(A_j | x_i) \qquad (2.6)$$

$$T(A_j | X)=P(X | A_j)T(A_j) / P(X),\ T(A_j) = 1 / \max(P(X | A_j) / P(X)) \quad (2.7)$$

The two formulas are asymmetrical because there is a statistical probability and a logical probability. $T(A_j)$ in (2.7) may be call longitudinally normalizing constant. Fig. 1 shows the relations between $T(A_j|X)$, $P(X|A_j)$, and $P(X)$.

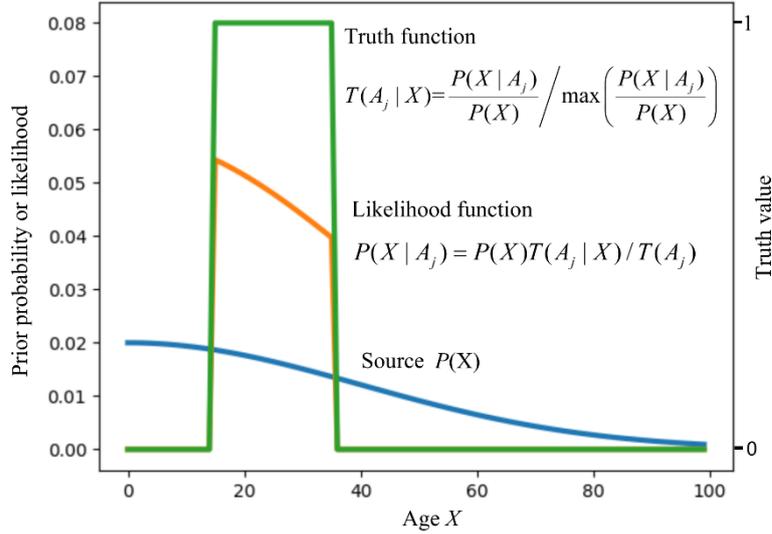

**Fig. 1** Relations between $T(A_j|X)$, $P(X|A_j)$ and $P(X)$ according to Bayes' Theorem III

If we follow Bayes' Theorem I or II, the normalizing constant in (2.7) should be $P(X)=\sum_i P(X)T(A_j|X)$ instead of $T(A_j)$. However, logical probability is not normalized, If we use $P(X)=\sum_i P(X)T(A_j|X)$, $P(X)$ is also not normalized. That is incorrect.

**The Proof of Bayes' Theorem III**: Since $P(X=\text{an}y, X \in A_j)= P(X|A_j)T(A_j) = T(A_j|X)P(X)$,

$$P(X | A_j) = P(X)T(A_j | X) / T(A_j), \quad T(A_j|X) = T(A_j)P(X | A_j) / P(X) \quad (2.8)$$

Since $P(X|A_j)$ is normalized, $T(A_j)=\sum_i P(x_i) T(A_j|x_i)$. Assume $x_j^*$ makes $P(x_j^*|A_j)/P(x_j^*)$ be the maximum of $P(X|A_j)/P(X)$. Since the maximum of $T(A_j|X)$ is 1,

$$T(A_j|x_j^*)=1=T(A_j)P(x_j^*|A_j)/P(x_j^*)$$

$$T(A_j)= 1/[P(x_j^*|A_j)/P(x_j^*)]=1/\max(P(X|A_j)/P(X)) \quad \textbf{QED.}$$

The second formula of Bayes' theorem III may be written as:

$$T(A_j | X)= \frac{P(X | A_j)}{P(X)} \bigg/ \max\left(\frac{P(X | A_j)}{P(X)}\right) \quad (2.9)$$

## 2.3 From Shannon's Channel to Semantic Channel

In Shannon's information theory [5], $P(X)$ is called the source and $P(Y)$ is called the destination, the transition probability matrix $P(Y|X)$ is called the channel. A Shannon's channel is formed of a group of transition probability functions:

$$P(Y|X): P(y_j|X), \ j=1, 2, …, n$$

Note that $P(y_j|X)$ is different from $P(Y|x_i)$; $P(y_j|X)$ ($y_j$ is constant and $X$ is variable) is also not normalized. $P(y_j|X)$ has two important properties: 1) It can be used for Bayes' prediction to get $P(X|y_j)$; after $P(X)$ becomes $P'(X)$, $P(y_j|X)$ still works; 2) $P(y_j|X)$ by a constant $k$ can make the same prediction because

$$\frac{P'(X)kP(y_j|X)}{\sum_i P'(x_i)kP(y_j|x_i)} = \frac{P'(X)P(y_j|X)}{\sum_i P'(x_i)P(y_j|x_i)} = P'(X|y_j) \quad (2.10)$$

The average of $\log[P(X|y_j)/P(X)]$ is Shannon's mutual information.

When $X=x_i$, $y_j(X)$ becomes a proposition $y_j(x_i)$, whose true value is $T(\theta_j|x_i)$. Similarly, a semantic channel is formed of a group of truth functions:

$$T(\theta|X): T(\theta_j|X), \ j=1, 2, …, n$$

Bayes' Theorem III to fuzzy sets is also tenable. According to (2.10), if $T(\theta_j|X) \propto P(y_j|X)$ or $T(\theta_j|X)=P(y_j|X)/\max(P(y_j|X))$, there is $P(X|\theta_j)=P(X|y_j)$.

## 2.4 Explaining Semantic Likelihood Function by GPS Positioning

Consider the semantic meaning of the small circle in a GPS map (see Fig. 2). The circle tells the position of the GPS device. A clock, a balance, or a thermometer is similar to a GPS device in that their actions may be abstracted as $y_j$="$X \approx x_j$". The $Y$ with such a meaning may be called an unbiased estimate, and its transition probability functions $P(y_j|X)$ constitute a Shannon channel, which may be expressed by

$$T(\theta_j|X)=\exp[-|X-x_j|^2/(2d^2)], j=1, 2, …, n \quad (2.11)$$

where $d$ is the standard deviation. The semantic likelihood function is

$$P(X|\theta_j) = \frac{P(X)\exp[-(X-x_j)^2/(2d^2)]}{\sum_i P(X)\exp[-(X-x_j)^2/(2d^2)]} \quad (2.12)$$

Fig. 2 (a) shows that a GPS device is used in a car; the positioning circle is on a building; the left side of the building is a highway and the right side is a road. We need to determine the most possible position of the car. If we think that the circle provides a likelihood function, we should infer "The car is most possibly on the building". However, common sense must deny this conclusion. It is easy to find that the semantic likelihood function in (2.12) accords with common sense.

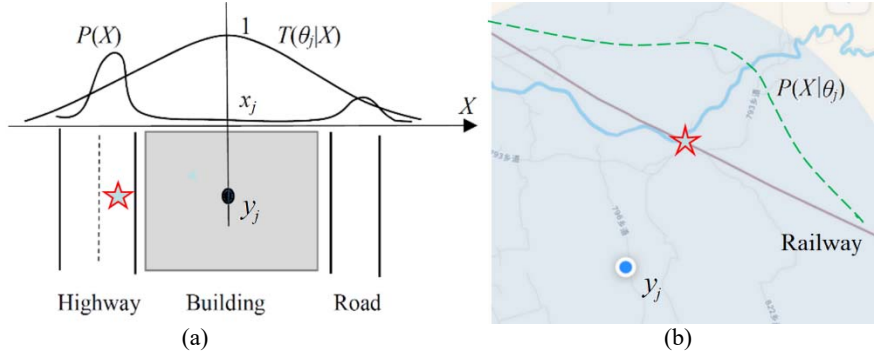

**Fig. 2.** Illustrating how semantic likelihood function accords with common sense by GPS's positioning with deviation. The (a) shows that the user drives a car; *P(X)* is changing. The (b) shows that the user is in a high-speed train; *P(X)* is not 0 only on a line.

Fig. 2 (b) shows an inaccurate positioning of a GPS device on a high-speed train. According to (2.12) or common sense, the most possible position is that marked by the star instead of the circle. However, by likelihood method or Bayesian inference, it is not easy to obtain this conclusion.

In natural language, many predictions are similar to GPS positioning and can be abstracted as "*X* is about $x_j$". The predictions provide truth functions instead of likelihood functions so that we can make semantic probability prediction $P(X|\theta_j)$.

## 3 Semantic Information Measure and Semantic Communication Optimization

### 3.1 Defining Semantic Information with Log Normalized Likelihood

The (amount of) semantic information conveyed by $y_j$ about $x_i$ is defined with [12]:

$$I(x_i;\theta_j) = \log \frac{P(x_i|\theta_j)}{P(x_i)} = \log \frac{T(\theta_j|x_i)}{T(\theta_j)} \qquad (3.1)$$

For an unbiased estimation $y_j$, its truth function may be that in (2.11). Hence $I(x_i;\theta_j)$ changes with $x_i$ as shown in Fig. 3.

This information criterion reflects Popper's thought [18]. It ensures that the larger the deviation is, the less information there is; the less the logical probability is, the more information there is; and, a wrong estimation conveys negative information.

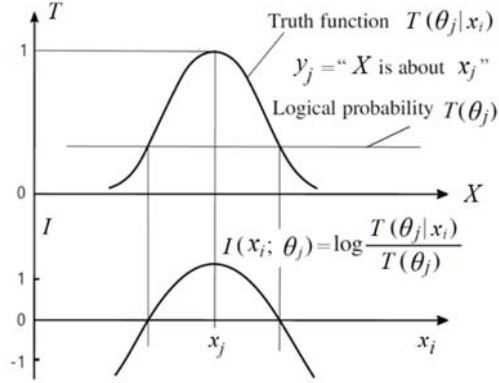

**Fig. 3.** Illustration of semantic information measure.

To average $I(x_i; \theta_j)$, we have [12]

$$I(X;\theta_j) = \sum_i P(x_i|y_j)\log\frac{P(x_i|\theta_j)}{P(x_i)} = \sum_i P(x_i|y_j)\log\frac{T(\theta_j|x_i)}{T(\theta_j)} \quad (3.2)$$

$$I(X;\theta) = \sum_j P(y_j)\sum_i P(x_i|y_j)\log\frac{P(x_i|\theta_j)}{P(x_i)}$$
$$= \sum_j\sum_i P(x_i,y_j)\log\frac{T(\theta_j|x_i)}{T(\theta_j)} = H(\theta) - H(\theta|X) \quad (3.3)$$
$$H(\theta) = -\sum_j P(y_j)\log T(\theta_j),\ H(\theta|X) = -\sum_j\sum_i P(x_i,y_j)\log T(\theta_j|x_i)$$

where $I(X; \theta_j)$ is generalized Kullback-Leibler information, and $I(X; \theta)$ is the semantic mutual information (a mutual cross entropy). It is easy to find that when $P(x_i|\theta_j)=P(x_i|y_j)$ for all $i, j$, $I(X; \theta)$ reaches its upper limit: Shannon's mutual information $I(X; Y)$. To bring (2.11) into (3.3), we have

$$I(X;\theta) = H(\theta) - H(\theta|X)$$
$$= -\sum_j P(y_j)\log T(\theta_j) - \sum_j\sum_i P(x_i,y_j)(x_i - x_j)^2/(2d_j^2) \quad (3.4)$$

It is easy to find that the maximum semantic mutual information criterion is similar to the regularized least squares criterion. $H(\theta|X)$ is similar to mean squared error and $H(\theta)$ is similar to negative regularization term.

Assume that there is a conditional sample $\{x(1), x(2), \ldots, x(N_j)\}$ for given $y_j$, and the sample points come from independently and identically distributed random variables. If $N_j$ is big enough, then $P(x_i|y_j)= N_{ij}/N_j$, where $N_{ij}$ is the number of $x_i$ in the sample. Then we have the log normalized likelihood:

$$\log \prod_i \left[\frac{P(x_i|\theta_j)}{P(x_i)}\right]^{N_{ji}} = N_j \sum_i P(x_i | y_j) \log \frac{P(x_i|\theta_j)}{P(x_i)} = N_j I(X;\theta_j) \quad (3.5)$$

Since $P(X)$ is irrelevant to $\theta_j$, the maximum semantic information criterion is equivalent to the maximum likelihood criterion. To average $I(X; \theta_j)$, there is "Average log normalized likelihood=Semantic mutual information $I(X; \theta)$". It is easy to prove that Shannon's mutual information $I(X; Y)$ is the upper limit of $I(X; \theta)$.

## 3.2 Receivers' Learning from Samples——Semantic Channel matches Shannon Channel

From the view point of semantic communication, the classification of the sender and that of the receiver are different. The receiver learns from samples to obtain labels' semantic meanings, i. e. the truth functions. The learning means logical classification. Then, when he receives $y_j$, he can predict $X$ to obtain $P(X|\theta_j)$ according to $P(X)$ and $T(\theta_j|X)$ so that he can make a decision. However, for a given $X$, the sender needs to select a label with most semantic information. This is selective classification. We may say that the logical classification is for the denotations of labels and selective classification is for the connotations of labels; label learning is letting a semantic channel match a Shannon's channel whereas label selecting is letting a Shannon's channel match a semantic channel.

When a sample is very big so that we can get the Shannon channel $P(Y|X)$ from the sample, by Bayes' theorem III and (2.12), we can directly obtain the optimized semantic channel:

$$T^*(\theta_j | X) = \frac{P(X|y_j)}{P(X)} \Big/ \max\left(\frac{P(X|y_j)}{P(X)}\right) = P(y_j|X) / \max(P(y_j | X)) \quad (3.6)$$

$T^*(\theta_j|X)$ does not changes with $P(X)$. If a sample is not big enough, we may use some parameters to construct $T(\theta_j|X)$, and train it by the sampling distribution $P(X|y_j)$:

$$T^*(\theta_j | X) = \arg\max_{T(\theta_j|X)} I(X;\theta_j) = \arg\max_{T(\theta_j|X)} \sum_i P(x_i | y_j) \log \frac{T(\theta_j | x_i)}{T(\theta_j)} \quad (3.7)$$

When samples are very big, $T^*(\theta_j|X)$ from (3.7) is the same as that from (3.6).

This learning method can make multi-label classifications much easier [19] because a multi-label classification can be naturally split into several single label classifications. If there are some examples for the negative hypothesis $y_j'$ of $y_j$, we need to optimize $T(\theta_j|X)$ by both positive and negative examples by

$$\begin{aligned} T^*(\theta_j | X) &= \arg\max_{T(\theta_j|X)} [I(X;\theta_j) + I(X;\theta_j^c)] \\ &= \arg\max_{T(\theta_j|X)} \sum_i \left[P(x_i | y_j) \log \frac{T(\theta_j | x_i)}{T(\theta_j)} + P(x_i | y_j') \log \frac{T(\theta_j^c | x_i)}{T(\theta_j^c)}\right] \end{aligned} \quad (3.8)$$

where $T(\theta_j^c|x_i)=1-T(\theta_j|x_i)$. $T^*(\theta_j|x_i)$ is only affected by $P(y_j|X)$ and $P(y_{j'}|X)$. Although those examples without $y_j$ or $y_{j'}$ affect $T^*(\theta_j)$, they do not affect $T^*(\theta_j|x_i)$. For a given label, this method actually divides all examples into three kinds: the positive, the negative, and the unclear. $T^*(\theta_j|x_i)$ is not affected by unclear instances.

If a negative label $y_j'$ does not appear in $D$, the second part will be 0. So, this binary classification is different from popular One vs Rest [3] classification, with which the problem is that an example $(x_i, y_1)$ without label $y_2$ does not means that $x_i$ makes $y_2'$ true. For example, $x_i$=25(age), $x_i$ makes both $y_1$="youth" and $y_2$="adult" true.

This binary logical classification allows the second part to be 0. It is a different from One vs Rest classification [4]. The new method for multi-label classification will be discussed in detail elsewhere [19]

### 3.4 Senders' Selecting Hypotheses or labels——Shannon's Channel matches Semantic Channel

For a seen instance $X$, the label sender selects $y_j$ by the classifier

$$y_j^* = h(X) = \arg\max_{y_j} \log I(\theta_j; x_i) = \arg\max_{y_j} \log \frac{T(\theta_j|X)}{T(\theta_j)} \qquad (3.9)$$

This classifier produces a noiseless Shannon channel. $T(\theta_j)$ happens to overcome the class-imbalance problem [4]. If $T(\theta_j|X)\in \{0, 1\}$, the information measure becomes Bar-Hillel and Carnap's information measure [6]; the classifier becomes

$$y_j^* = h(X) = \arg\max_{y_j, T(A_j|X)=1} \log[1/T(A_j)] = \arg\min_{y_j, T(A_j|X)=1} T(A_j) \qquad (3.10)$$

It means that we should select a label with the least logical probability or with the richest connotation as Popper pointed out [18].

Although $T^*(\theta_j|X)$ does not change with $P(X)$, the classifier $h(X)$ does. Assume that $y_j$="Old person", $T^*(\theta_j|X)=1/[1+\exp(-0.2(X-75))]$, and $P(X)=1-1/[1+\exp(-0.15(X-c))]$. Fig. 4 shows that the optimized dividing point $z^*$ changes with $c$.

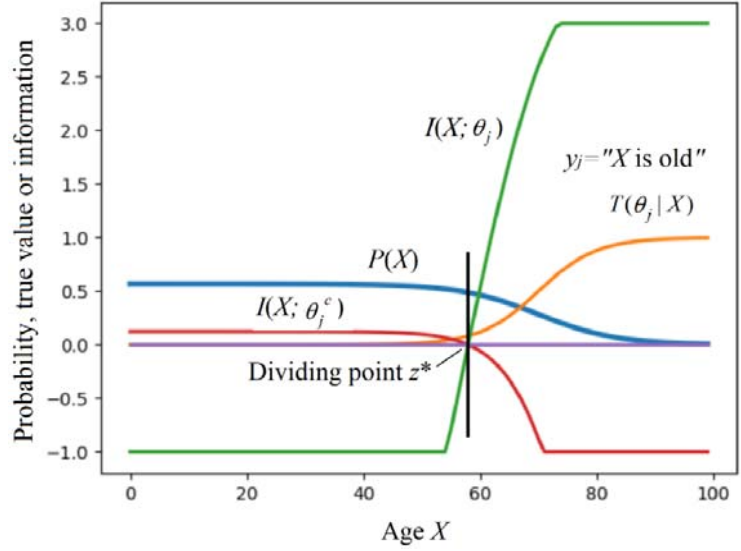

Fig. 4. The optimized dividing point $z^*$ for label "Old person", which changes with $P(X)$.

**Table 1** The classifier $h(X)$ for $y_j$="Old person" changes with $P(X)$

| c | Population density decreasing area | Dividing point |
|---|---|---|
| 50 | 40-60 | $y_j=h(X>48)$ |
| 60 | 50-70 | $y_j=h(X>54)$ |
| 70 | 60-80 | $y_j=h(x>57)$ |

The dividing point of $h(X)$ moves rightward when old population increases because the semantic information criterion with class-imbalance consideration encourages us to reduce the failure of reporting small probability events. Longevous population's increasing can change $h(X)$; new $h(X)$ will change Shannon's channel and produce new sample; new semantic channel will match new Shannon's channel… The semantic meaning of "Old person" has been evolving with human lifetimes in this way.

## 4 Semi-Supervised Learning and Non-Supervised Learning

### 4.1 Medical Test and Confirmation Measure

To a medical test or an uncertain hypothesis, we may believe it to different degree. After optimizing the degree of belief $b$ by statistical data, we can obtain the best degree of belief $b^*$ with most semantic information. The $b^*$ can be called a confirmation measure [20].

For a medical test (shown in Fig. 5), $U=\{x_0, x_1\}$ where $x_0$ means an uninfected person and $x_1$ means an infected person, and $V=\{y_0, y_1\}$ where $y_0$ means test-negative and $y_1$ means test-positive.

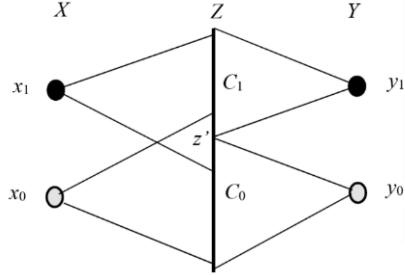

Fig. 5. Illustrating a medical test. It can be abstracted as a 2×2 Shannon noisy channel and the Shannon mutual information changes with partition point $z'$.

In medical tests, the conditional probability in which a test for an infected testee is positive is called sensitivity, and the conditional probability in which a test for an uninfected testee is negative is called specificity [21]. The sensitivity and specificity form a Shannon channel as shown in Table 2.

**Table 2**. The sensitivity and specificity of medical tests form a Shannon's channel $P(Y|X)$

| $Y$ | Infected $x_1$ | Uninfected $x_0$ |
|---|---|---|
| Positive $y_1$ | $P(y_1|x_1)$=sensitivity | $P(y_1|x_0)$=1-specificity |
| Negative $y_0$ | $P(y_0|x_1)$=1-sensitivity | $P(y_0|x_0)$=specificity |

If we absolutely believe that a test-positive means being infected, and a test-negative means not being infected, then there are truth values $T(y_1|x_1)=T(y_0|x_0)=1$, $T(y_1|e_0)=T(y_0|x_1)=0$. If we use these truth values as the semantic channel, the semantic information will be negatively infinite when one counterexample exists. Thus, we need to consider the degree of belief of $y_j$. Let it be denoted by $b_j$. Let the degree of disbelief be denoted by $b_j'=1-|b_j|$. Then the truth function of $y_j$ may be defined as [22]

$$T(\theta_j|X) = b_j' + b_j T(y_j|X) \qquad (4.1)$$

Here, $b'$ is actually the truth value of a counterexample. Then Table 3 shows the semantic channel for medical tests.

**Table 3.** Two degrees of disbelief of a medical test form a semantic channel $T(\theta|X)$

| $Y$ | Infected $x_1$ | Uninfected $x_0$ |
|---|---|---|
| Positive $y_1$ | $T(\theta_1|x_1)=1$ | $T(\theta_1|x_0)=b_1'$ |
| Negative $y_0$ | $T(\theta_0|x_1)=b_0'$ | $T(\theta_0|x_0)=1$ |

According to (3.6), two confirmation measures are

$$b_1'^* = P(y_1|x_0)/P(y_1|x_1); \quad b_0'^* = P(y_0|x_1)/P(y_0|x_0) \quad (4.2)$$

In the medical community, Likelihood Ratio is used to indicate how good a test is [23]. The confirmation measure is compatible with Likelihood Ratio measure. There are

$$LR^+ = P(y_1|x_1)/P(y_1|x_0) = 1/b_1'^*; \quad LR^- = P(y_0|x_0)/P(y_0|x_1) = 1/b_0'^* \quad (4.3)$$

The LR has been used by Thornbury et al for probability prediction [21]. Still, it is easier to use $b_1'^*$ and $b_0'^*$ for probability prediction [22], such as $P'(X|\theta_1) = P'(X)/(P'(x_1) + b_1'^* P'(x_0))$.

Assume that $Np$ is the number of positive examples; $Nc$ is the number of counter-examples; $Np+Nc$ is big enough. The $b^*$ may also be negative [22]. There is

$$b^* = \begin{cases} 1 - Nc/Np, & Nc < Np \\ Np/Nc - 1, & Nc \geq Np \end{cases} \quad (4.3)$$

For example, when $Nc/Np=2$, $b_1^*=1/2-1=-0.5$, which means the hypothesis is misleading. This confirmation measure is simillar to and yet different from Confidence Level (CL). When $Nc<Np$,

$$b^* = 1 - Nc/Pc; \quad CL = Np/(Np+Nc) = 1/(2-b^*) \quad (4.4)$$

When $Nc=Np$, the hypothesis should be meaningless. We have $b^*=0$ and $CL=0.5$. When $Nc>Np$, the hypothesis is misleading. We have $-1 \leq b^* < 0$ and $0 < CL < 0.5$. It is obvious that the confirmation measure $b^*$ reflects the support from statistical data better.

### 4.2 The CM algorithm for Estimations and Unseen Instance Classifications

For tests, discrete estimations, and unseen instance classifications, we need the average semantic information criterion. Assume that $Z$ is an observed datum; $Z \in C = \{z_1, z_2, ...\}$; $X_L \in U_L = \{X_1, X_2, ...\}$ is a true class or true label. If for given $Z=z_k$, the predicted distribution of $X$ is $P(X_L|z_k)$, we may select the optimized label

$$y_j^* = \arg\max_{y_j} I(X_L; \theta_j|z_k) = \arg\max_{y_j} \sum_i P(X_i|z_k) \log \frac{T(\theta_j|X_i)}{T(\theta_j)} \quad (4.5)$$

This classifier produces a noisy Shannon's channel. However, $P(Y|X_L)$ so obtained is not optimal. We need an iterative algorithm for this semi-supervised learning.

Let $C_j$ be a subset of $C$ and $y_j=f(Z|Z \in C_j)$. Hence $S=\{C_1, C_2, ...\}$ is a partition of $C$. Our aim is to find optimal $S$, which is

$$S^* = \arg\max_S I(X_L; \theta|S) = \arg\max_S \sum_j \sum_i P(C_j) P(X_i|C_j) \log \frac{T(\theta_j|X_i)}{T(\theta_j)} \quad (4.6)$$

**Matching I**: Let the semantic channel match the semantic channel. From given $P(X_L, Z)$ and $S$, we can obtain the Shannon channel:

$$P(y_j | X_i) = \sum_{z_k \in C_j} P(z_k | X_i), \ i=1,2,...,n; j=1,2,...,n \qquad (4.7)$$

Then we obtain the semantic channel $T(\theta|X)$ according to (3.7).

**Matching II**: Let the Shannon channel match the semantic channel by

$$P(y_j | Z) = \lim_{s \to \infty} \frac{[\exp(I(X_L; \theta_j | Z))]^s}{\sum_{j'} [\exp(I(X_L; \theta_{j'} | Z))]^s}, \quad j=1, 2, ..., n \qquad (4.8)$$

Since $P(y_j|Z)=0$ or 1 when $s \to \infty$, the above formula provides a classifier $Y=f(Z)$ and a new $S$. Repeating Matching I and Matching II until $S$ does not change. The convergent $S$ is the $S^*$ we seek.

We use two examples to show the CM iteration algorithm for a test and a discrete estimation. Assume that $Z \in C=\{1, 2, ..., 100\}$ and $P(Z|X_L)$ is a group of Gaussian distributions (where $K_i$ is a normalizing constant):

$$P(Z|X_i)=K_i \exp[-(Z-c_i)^2/(2d_i^2)], \ i=0, 1, ...$$

**Example 1** Find $S^*$ for a test (with a 2*2 Shannon's channel).

Input data: $P(x_0)=0.8$; $c_0=30$, $c_1=70$; $d_0=15$, $d_1=10$. The start dividing point $z'=50$.

The iterative process: Matching II-1 gets $z'=53$; Matching II-2 gets $z'=54$; Matching II-3 gets $z^*=54$.

**Example 2** Find $S^*$ for a discrete estimation with a 3*3 Shannon's channel. A pair of good start dividing points and a pair of bad start dividing points (shown in Fig. 6) are used to check the convergence and speed of the iteration.

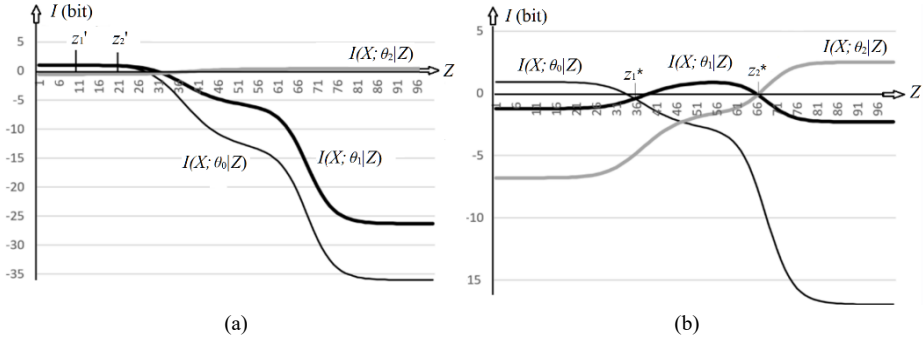

Fig. 6 The iteration with bad start dividing points. At the beginning of the iteration (a), three information curves cover very small positive areas. At the end of the iteration (b), three information curves cover much larger positive areas so that $I(X; \theta)$ reaches its maximum.

Input data: $P(X_0)=0.5$, $P(x_1)=0.35$, and $P(x_2)=0.15$; $c_0=20$, $c_1=50$, and $c_2=80$; $d_0=15$, $d_1=10$, and $d_2=10$.

The iterative results: a) With the good start dividing points: $z_1'=50$ and $z_2'=60$, the number of iterations is 4; $z_1^*=35$ and $z_2^*=66$. b) With the bad start dividing points:

$z_1$'=9 and $z_2$'=20, the number of iterations is 11; $z_1$*=35 and $z_2$*=66 also. Fig. 4 shows the information curves over $Z$ before and after the iteration.

This iterative example shows that the CM algorithm is fast and reliable. The convergence can be proved with the help of the $R(G)$ function [14].

**4.2 Non-Supervised Learning: the CM Algorithm for Mixture Models**

Assume a sampling distribution $P(X)$ is produced by the conditional probability $P^*(X|Y)$ being some function such as Gaussian distribution. We only know the number of the mixture components, without knowing $P(Y)$. We need to solve $P(Y)$ and model parameters $\theta$, so that the predicted distribution of $X$, denoted by $Q(X)$, is as close to $P(X)$ as possible, i. e. the relative entropy $H(Q||P)= -\sum_i P(x_i)\log Q(x_i)$ is as small as possible.

With the CM algorithm, we can improve the EM algorithm [23] or VBEM algorithm [24] as follows:

**Left-step a:** Construct Shannon's channel by

$$P(y_j | X) = P(y_j)P(X|\theta_j)/Q(X), \quad Q(X) = \sum_j P(y_j)(X|\theta_j), \quad j=1, 2, ..., n \quad (4.9)$$

This formula has been used in the EM algorithm [23].

**Left-step b** Use the following equation to obtain a new $P(Y)$ repeatedly until the inner iteration converges:

$$P(y_j) \Leftarrow \sum_i P(x_i)P(y_j|x_i) = \sum_i P(x_i)\frac{P(x_i|\theta_j)}{\sum_k P(y_k)P(x_i|\theta_k)}P(y_j), \quad j=1, 2, ..., n \quad (4.10)$$

If $H(Q||P)$ is less than a small number, such as 0.001 bit, then end the iteration.

**Right-step:** Optimize the parameters in the likelihood function $P(X|\theta)$ on the right of the log in (4.11) to maximize the semantic mutual information:

$$I(X;\theta) = \sum_i \sum_j P(x_i)\frac{P(x_i|\theta_j)}{Q(x_i)}P(y_j)\log\frac{P(x_i|\theta_j)}{P(x_i)} \quad (4.11)$$

Then go to Left-step a.

Fortunately, to prove $H(Q||P)\rightarrow 0$, we derived an important formula [16]

$$\min H(Q||P) = \min_{P(Y),\theta}(I(X;Y)-I(X;\theta))= \min_{P(Y),\theta}(R(G)-G) \quad (4.12)$$

where $G$ is the semantic information and $R(G)$ is the minimum Shannon's mutual information for given $G$. In every step, $H(Q||P)$ is decreasing. In comparison with the EM algorithm, the CM algorithm has faster speed and clearer convergence reason [16].

Table 2 shows the parameters and iterative results for $R>R^*=H(X)-H^*(X|Y)$. The iterative process is shown in Fig. 7.

**Table 2.** Real and guessed model parameters and iterative results for Example 2 ($R>R^*$)

| Y | Real parameters | | | Start parameters $H(Q||P)$=0.680 bit | | | Parameters after 5 iterations $H(Q||P)$=0.00092 bit | | |
|---|---|---|---|---|---|---|---|---|---|
| | c | d | $P^*(Y)$ | c | d | $P(Y)$ | c | d | $P(Y)$ |
| $y_1$ | 35 | 8 | 0.1 | 30 | 8 | 0.5 | 38 | 9.3 | 0.134 |
| $y_2$ | 65 | 12 | 0.9 | 70 | 8 | 0.5 | 65.8 | 11.5 | 0.886 |

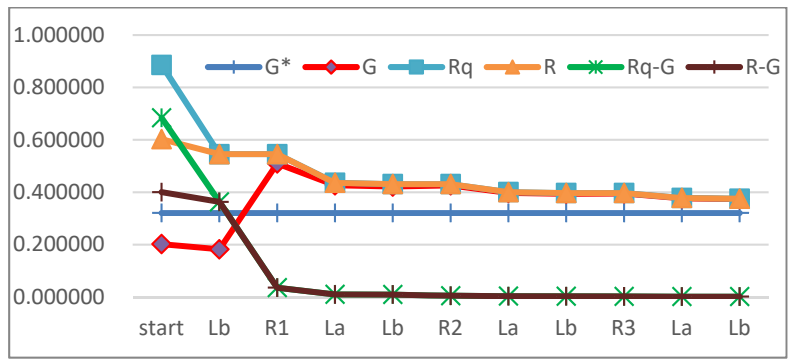

**Fig. 7.** The iterative process as $R>R^*$. $Rq$ is $R$ before Step Left b. Eq. (15). $H(Q||P)$ decreases in all steps. $R$ is monotonically decreasing. $G$ increases more or less in all Right-steps and decreases in all Left-steps. $G$ and $R$ gradually approach $G^*=R^*$ so that $H(Q||P)=R-G$ is close to 0.

This example is a challenge to all authors who prove that the standard EM algorithm or a variant EM algorithm converges [23-25] by that $\log P(X^N, Y|\Theta)$ or other likelihood is monotonically increasing or no-decreasing in all steps. In Example 2, $Q^*=-NH^*(X, Y)=-6.031N$. After the first optimization of parameters, $Q=-6.011N>Q^*$. If we continuously maximize $Q$, $Q$ cannot approach less $Q^*$.

## 4   Discussions and Summary

This paper introduces the mutual matching of semantic channel and Shannon' channel via the third kind of Bayes' theorem and the semantic information method for machine learning, including semi-supervised learning and non-supervised learning.

In comparison with Bayesian inference [26], the semantic information method is more compatible with traditional Bayes' prediction for $P(X|y_j)$ with $P(y_j|X)$ and $P(X)$.

Regularized Least Squares criterion is getting popular. It seems that likelihood criterion is out of date. However, this paper shows that the semantic information criterion is compatible with both.

The mutual matching of semantic channel and Shannon's channel is very similar to the mutual contest of generator and discriminator in GAN [27]. The relationship between the two methods is worth analyzing.

The rationality of the semantic information method is supported by: 1) It can explain better how we obtain the semantic meanings of hypotheses or labels, how we make hypotheses according to the semantic meanings and the source, and how the semantic meaning was evolving; 2) It is concise and suitable to different types of learning; 3) It is compatible with the thoughts of Bayes, Fisher, Shannon, Zadeh, Tarski, Davidson, Wittgenstein, and Popper.

It is expected that the semantic information method will be a competitive alternative to Bayesian inference.